\documentclass[12pt,journal,compsoc,onecolumn]{IEEEtran}

\usepackage{amsmath,graphicx}
\usepackage{amssymb,amsfonts}
\usepackage{dblfloatfix}
\usepackage{multirow}
\usepackage[noadjust]{cite}

\usepackage[ruled,vlined]{algorithm2e}
\usepackage{graphicx}
\newcommand{\cmark}{\ding{51}}
\newcommand{\xmark}{\ding{55}}
\usepackage{pifont}

\usepackage{booktabs}
\usepackage{siunitx}
\usepackage[numbers,compress]{natbib}
\usepackage{texnames}
\usepackage{bm,bbm}

\usepackage{lipsum}

\def\HH{{\bf H}}
\def\tr{{\bf tr}}

\def\XX{{\bf X}}

\def\D{{d}}

\def\tr{{\bf tr}}
\def\1{{\bf 1}}

\def\V{{\bf V}}

\def\b1{{\bf 1}}

\usepackage{amsthm}
\newtheorem{proposition}{Proposition}
\newtheorem{definition}{Definition}

\usepackage[ruled,vlined]{algorithm2e}

\newlength{\bibitemsep}
\newlength{\bibparskip}
\let\oldthebibliography\thebibliography
\renewcommand\thebibliography[1]{
  \oldthebibliography{#1}
  \setlength{\parskip}{\bibitemsep}
  \setlength{\itemsep}{\bibparskip}
}

\title{Active Learning for GCN-based Action Recognition}
\author{Hichem Sahbi \\
$ $ \\
Sorbonne University, CNRS, LIP6,  F-75005, Paris, France 
 }

\begin{document}
 \maketitle
\begin{abstract}
  Despite the notable success of graph convolutional networks (GCNs) in skeleton-based action recognition, their performance often depends on large volumes of labeled data, which are frequently scarce in practical settings. To address this limitation, we propose a novel label-efficient GCN model. Our work makes two primary contributions. First, we develop a novel acquisition function that employs an adversarial strategy to identify a compact set of informative exemplars for labeling. This selection process balances representativeness, diversity, and uncertainty. Second, we introduce bidirectional and stable GCN architectures. These enhanced networks facilitate a more effective mapping between the ambient and latent data spaces, enabling a better understanding of the learned exemplar distribution. Extensive evaluations on two challenging skeleton-based action recognition benchmarks reveal significant improvements achieved by our label-efficient GCNs compared to prior work.
\end{abstract}

\section{Introduction}Skeleton-based action recognition involves the analysis of articulated human body configurations through the extraction of skeletal joint coordinates and the modeling of their spatio-temporal relationships. Early approaches relied on the design of handcrafted features~\cite{Yun2012,refref18,Ji2014,Li2015a,refref40,refref41,refref59,refref39,aaaaa3}, such as inter-joint angles and relative Euclidean distances, which were subsequently employed as input to classification algorithms including support vector machines and hidden Markov models~\cite{Sahbiling2013,Vemulapalli2014,refref11}, or integrated with manifold learning techniques~\cite{RiemannianManifoldTraject,refref61,Huangcc2017,ref23}. The resurgence of deep learning~\cite{krizhevsky2012imagenet,refref10,pose3D} led to the widespread adoption of recurrent neural networks, particularly Long Short-Term Memory (LSTM) and Gated Recurrent Unit (GRU) architectures~\cite{Du2015,Liu2016,Zhua2016,Zhang2017,Zhang2017b,GCALSTM,DeepGRU}, for their capacity to model the temporal evolution inherent in skeletal sequences. More recently, Graph Convolutional Networks (GCNs) have gained prominence, exploiting the intrinsic graph topology of human skeletons to learn spatial correlations between joints. Furthermore, attention mechanisms integrated with GCNs~\cite{Jiang2020,Song2017,Liu2021,plizzari2023spatio,refffabc888,sahbicbmi24,sahbi2021kernel} have demonstrated substantial performance enhancements by effectively capturing long-range dependencies and complex kinematic patterns.\\

\indent The performance of learning-based approaches in skeleton-based action recognition is fundamentally constrained by the availability of large, diverse datasets with high-fidelity skeletal sequence annotations. Acquiring such datasets represents a significant bottleneck, demanding substantial time and manual effort. To alleviate data and label scarcity, various strategies have been proposed. Data augmentation \cite{DA2} artificially expands dataset size and variability. Few-shot and transfer learning \cite{TL1} exploit knowledge from related domains. Self-supervised learning \cite{SS1} aims to learn intrinsic data representations without explicit labels. However, while these knowledge-leveraging strategies offer valuable benefits, their effectiveness in bridging the accuracy gap inherent in limited direct supervision is often predicated on the relevance of the derived knowledge. Ultimately, despite the importance of leveraging existing knowledge, the quality and relevance of directly annotated data remain the primary and often most impactful determinants of achieving optimal model performance. \\

\indent In contrast to the aforementioned passive learning paradigms, active learning \cite{Burr2009,refff33333} offers a more resource-efficient and focused methodology for dataset construction. By adequately selecting the most informative instances for annotation, active learning maximizes the learning potential of a model while minimizing the requisite human labeling effort. This iterative process of querying an oracle (human annotator) for labels on samples exhibiting the highest uncertainty or representativeness prioritizes the acquisition of data most likely to yield significant gains in model accuracy. Consequently, active learning not only alleviates the overall labeling burden but also ensures the resulting labeled dataset is optimally aligned with the specific recognition task. Particularly in contexts where data or label acquisition incurs high costs or time investments, active learning presents a compelling alternative by directly addressing the fundamental need for high-quality and task-relevant labeled data.\\

\indent The selection of informative data within active learning aims to identify samples that maximally enhance a model's learning capacity \cite{sahbi2023aactive}. Sophisticated strategies, including query-by-committee \cite{seung1992query}, expected model change maximization \cite{settles2010active}, and deep reinforcement learning \cite{ren2018learning,sahbiactivel1,sahbiactivel2}, have been developed to optimize the informativeness of selected samples. These methods typically integrate measures of uncertainty~\cite{SS1,Ref54,Ref56,Ref57,Culotta2005,kirsch2022bayesian} and diversity~\cite{Ref50a,Ref51a} within various application contexts~\cite{du2022efficient,cai2022self,kondo2023active,aljundi2023online,zhao2023active}. Uncertainty-based strategies, such as margin sampling and entropy-based criteria~\cite{zhao2023uncertainty,kim2023batch}, prioritize samples where the model exhibits low predictive confidence, thereby focusing subsequent training on areas of maximal ambiguity. Complementarily, diversity-based methods, including coverage maximization~\cite{wu2022adaptive,zhang2022multi} and core-set \cite{sener2018active}, aim to select a representative subset that spans the entirety of the data distribution, ensuring exposure to a wide spectrum of data variations. Representativeness-based approaches \cite{ashfaq2023clustering} further contribute by seeking samples that closely approximate the overall data distribution, fostering a balanced learning process. While these established criteria offer valuable heuristics, many current implementations lack a strong theoretical foundation. Future advancements should focus on developing selection criteria rigorously grounded in probabilistic frameworks to enable the identification of truly optimal and informative subsets. Such principled approaches would not only enhance the efficiency of active learning but also provide a more robust methodology for constructing highly effective training datasets.\\

\indent Addressing the aforementioned limitations, this paper introduces a label-efficient GCN for skeleton-based action recognition. The core contribution of the proposed method lies in a novel, principled probabilistic framework that designs unlabeled exemplars (candidate samples for labeling) rather than passively selecting them from a static pool of unlabeled data. These exemplars are derived as an interpretable solution to a well-defined objective function that integrates data representativeness, diversity, and uncertainty. Our framework achieves this exemplar design through a novel, stable, and invertible bidirectional GCN. This architecture enables the mapping of input graphs, residing on highly nonlinear manifolds, from the ambient (input) space to a more tractable latent space. Notably, the proposed GCNs induce a standard probability distribution (specifically, a Gaussian) in the latent space, facilitating more efficient sampling and search for exemplars compared to the arbitrary distributions prevalent in the ambient space. Once identified, these learned exemplars are mapped back to the input space, leveraging the invertibility and stability of our GCNs. In essence, the proposed framework enables the design of bidirectional GCNs that exhibit both robust classification and effective exemplar design capabilities – even under data-frugal conditions – without the necessity of auxiliary generative networks. Comprehensive experiments, conducted on two challenging skeleton-based action recognition benchmarks, demonstrate the superior performance of our label-efficient method in comparison to related state-of-the-art approaches.
\section{Display Model}
Our proposed Active Learning (AL) framework comprises two principal components: {\it display} model and {\it learning} model. The display model implements an acquisition function to identify the most informative unlabeled data points, which are subsequently presented to an oracle for annotation. The learning model then retrains a label-efficient classifier using the newly acquired labels. These two stages are executed iteratively until a predefined classification accuracy target is met or a labeling budget is exhausted. Formally, let $\mathcal{U} = \{\mathbf{x}_1, \dots, \mathbf{x}_n\} \subset \mathbb{R}^p$ represent the pool of unlabeled data. At each AL iteration $t \in \{0, \dots, T-1\}$, the {\it display} model, detailed in Section~\ref{display}, constructs a subset $\mathcal{D}_t$---termed the display set---which is utilized to query the oracle for corresponding labels $\mathcal{Y}_t$. A classifier $f_t$ is subsequently trained on the incrementally expanded labeled dataset $\bigcup_{k=0}^{t} (\mathcal{D}_k, \mathcal{Y}_k)$. Our primary contribution, introduced in Section~\ref{display}, centers on a novel model that constructs displays in a {\it flexible manner} rather than sampling fixed subsets from $\mathcal{U}$. 
\subsection{Display model design}\label{display}
Our proposed method is adversarial and aims to select the most diverse, representative, and uncertain data points to effectively {\it challenge} the current classifier $f_t$, thereby facilitating the training of an enhanced classifier $f_{t+1}$ in the subsequent active learning iteration. Rather than directly sampling the display set $\mathcal{D}_{t+1}$ from the unlabeled pool $\mathcal{U}$, we employ a probabilistic framework to construct $\mathcal{D}_{t+1}$ (denoted as $\mathcal{D}$). Let $\mathbf{X} \in \mathbb{R}^{p \times n}$ and $\mathbf{V} \in \mathbb{R}^{p \times K}$ represent the matrices of the unlabeled pool $\mathcal{U}$ and the display set $\mathcal{D}$, respectively, where $K = |\mathcal{D}|$. To construct the display $\mathbf{V}$, we define a conditional probability distribution for each column $\mathbf{V}_k$, quantifying the membership (or contribution) $\mu_{ik}$ of each unlabeled data point $\mathbf{x}_i \in \mathcal{U}$ in the formation of $\mathbf{V}_k$. The memberships $\mu = \{\mu_{ik}\}_{ik}$ and the resulting display $\mathbf{V}$ are determined by minimizing the following constrained objective function
\begin{equation}\label{of} 
 \begin{array}{c} 
\displaystyle  \min_{\mu \in \Omega,\V}  \displaystyle \tr(\mu \ \D(\XX,\V)^\top)  + \alpha \ \sum_{k,k'}^{K,N}  \exp\bigg\{-\frac{1}{\sigma}\big\| \V_k - \HH_{k'}\big\|_2^2\bigg\} \\
           + \ \beta \ \tr(\V^\top \V)  + \gamma \  \tr(\mu^\top \log \mu),
\end{array}   
\end{equation}
\noindent where  $\Omega = \{\mu : \mu \geq 0, \mathbf{1}_n^\top \mu = \mathbf{1}_K^\top\}$  is a convex set enforcing $\mu$ to be column-stochastic (i.e., each column represents a conditional probability distribution), and $\mathbf{1}_K$ and $\mathbf{1}_n$ are vectors of $K$ and $n$ ones, respectively, with $^\top$ denoting the transpose. The objective function (Eq.~\ref{of}) comprises four weighted terms: the \textbf{Representativity} term minimizes the divergence between the designed exemplars in $\mathbf{V}$ and the data distribution of $\mathcal{U}$, ensuring that the oracle's annotations are based on realistic and representative exemplars, thereby mitigating the selection of trivial or semantically irrelevant data points. The \textbf{Diversity} term maximizes the dissimilarity between the $N$ previously selected exemplars (represented by the matrix $\mathbf{H}$) and the $K$ currently selected exemplars (matrix $\mathbf{V}$), enforcing the selection of new exemplars that are maximally distinct from the previously acquired labeled data. The \textbf{Uncertainty} term quantifies the predictive ambiguity associated with the exemplars in $\mathbf{V}$, encouraging the selection of exemplars situated near the decision boundaries of the learned classifiers, which is crucial for reducing model uncertainty and accelerating the convergence towards well-defined decision functions; it also serves as a regularizer on $\mathbf{V}$. Finally, the \textbf{Regularization of $\mu$} term promotes uniform conditional probabilities $\mu = \{\mu_{ik}\}_{ik}$ in the absence of strong prior information favoring specific data point contributions across the other three terms. The relative influence of these terms is controlled by the non-negative weights $\alpha, \beta, \gamma$, whose setting is discussed subsequently.
\begin{proposition}\it 
The optimality conditions of Eq.~\ref{of} yield the following iterative update of  the solution as a fixed point of
\begin{equation}\label{eq2}
  \begin{aligned}
\mathbf{\mu}^{(\tau+1)} &= \hat{\mathbf{\mu}}^{(\tau+1)} \mathbf{diag} \left( \mathbf{1}_n^\top \hat{\mathbf{\mu}}^{(\tau+1)} \right)^{-1} \\
\mathbf{V}^{(\tau+1)} &= \hat{\mathbf{V}}^{(\tau+1)} \left( \mathbf{diag} \left( \mathbf{1}_n^\top \mathbf{\mu}^{(\tau)} \right) + \beta \mathbf{I} \right)^{-1},
\end{aligned}
\end{equation}
where $\hat{\mathbf{\mu}}^{(\tau+1)}$ and $\hat{\mathbf{V}}^{(\tau+1)}$ are given by
\begin{equation}\label{eq3}
\begin{aligned}
\hat{\mathbf{\mu}}^{(\tau+1)} &= \exp\left\{-\frac{1}{\gamma}d(\mathbf{X},\mathbf{V}^{(\tau)})\right\} \\
\hat{\mathbf{V}}^{(\tau+1)} &= \mathbf{X} \mathbf{\mu}^{(\tau)} - \frac{2\alpha}{\sigma} \left( \mathbf{V}^{(\tau)} \mathbf{diag}(\mathbf{1}_N^\top \mathbf{S}) - \mathbf{H} \mathbf{S} \right),
\end{aligned}
\end{equation}
with the similarity matrix $\mathbf{S}$ between $\mathbf{V}$ and $\mathbf{H}$ (where $\mathbf{V}^{(\tau)}$ is denoted as $\mathbf{V}$ for brevity) defined as
\begin{equation}\label{eqqqabc}
\mathbf{S} = \exp\left\{-\frac{1}{\sigma} \left(\mathbf{1}_N \mathbf{diag}(\mathbf{V}^\top\mathbf{V})^\top + \mathbf{diag}(\mathbf{H}^\top\mathbf{H}) \mathbf{1}_K^\top - 2\mathbf{H}^\top\mathbf{V}\right)\right\},
\end{equation}
here $\mathbf{1}_N$ is a vector of $N$ ones, $\mathbf{1}_K$ is a vector of $K$ ones, and $\mathbf{diag}(\cdot)$ transforms a vector into a diagonal matrix.
\end{proposition}
In view of space limitation, details of the proof for the aforementioned iterative updates, stemming from the gradient of the objective function in Eq.~\ref{of}, are omitted. The solution for $\mathbf{\mu}$ in Eq.~\ref{eq3} notably demonstrates an inverse correlation between data point distances and their membership values: smaller distances between the input data $\mathbf{X}$ and the designed exemplars $\mathbf{V}$ result in higher membership values, and vice versa. Furthermore, each exemplar $\mathbf{V}_k$ is constructed as a combination of two components. The primary component is a normalized linear combination of the input data points, weighted by their memberships to $\mathbf{V}_k$. The secondary component, scaled by $\alpha$, disrupts $\mathbf{V}_k$ to maximize its dissimilarity from the previously selected exemplars in $\mathbf{H}$.
The iterative optimization process starts with random initializations for $\mathbf{\mu}^{(0)}$ and $\mathbf{V}^{(0)}$, and empirically converges to a near-optimal solution $(\tilde{\mathbf{\mu}}, \tilde{\mathbf{V}})$ within a few iterations. This converged solution determines the subsequent display set $\mathcal{D}_{t+1}$ used for training the classifier $f_{t+1}$. The parameters $\alpha$ and $\beta$ are set to balance the impact of their respective terms, specifically $\alpha = \frac{1}{KN}$ and $\beta = \frac{1}{Kp}$. In Eq.~\ref{eq3}, $\sigma$ is set proportionally to $\alpha$ to absorb the former by the latter. The hyperparameter $\gamma$, scaling the exponential in $\hat{\mathbf{\mu}}^{(\tau+1)}$, is dynamically adjusted per iteration based on the magnitude of its input, specifically $\gamma = \frac{1}{nK} \|\log(\hat{\mathbf{\mu}}^{(\tau+1)})\|_1$.\\

\noindent Considering the AL formulation detailed above, this paper explores two variants of our proposed solution. The first one directly identifies exemplars in the ambient (input) space using the derived formulation. The second variant, leveraging the invertibility and stability of our learned GCNs (as demonstrated in Section~\ref{learning}), identifies exemplars in the latent space and subsequently maps them back to the ambient space. As will be shown through experiments, performing exemplar design in the latent space via an invertible and stable GCN mapping yields a significant improvement in AL performance.
\section{Learning Model}\label{learning}
As previously established, the effectiveness of AL is critically dependent on the fidelity of the display model. Ideally, the generated displays should accurately represent the underlying data distribution in the input space. However, a significant limitation can arise when dealing with intricate, nonlinear distributions, potentially affecting the display model defined in Eq.~\ref{of}. Ensuring that the generated displays remain consistent with data residing on nonlinear manifolds presents a considerable challenge. Consequently, in the subsequent section, we revisit GCNs and introduce—as our second contribution—a novel learning model designed to overcome this limitation by training GCNs that are bidirectional, invertible, and stable.
\subsection{Graph convnets at a glance}
Let $\{\mathcal{G}_i=(\mathcal{V}_i, \mathcal{E}_i)\}_i$ denote a collection of graphs, where $\mathcal{V}_i$ and $\mathcal{E}_i$ represent the node and edge sets of  $\mathcal{G}_i$, respectively. For clarity, let us focus on a single graph $\mathcal{G} = (\mathcal{V}, \mathcal{E})$ from this set. Each node $v \in \mathcal{V}$ in $\mathcal{G}$ is associated with a signal $\psi(v) \in \mathbb{R}^s$, and the graph structure is defined by an adjacency matrix $\mathbf{A}$. GCNs aim to learn a set of $C$ filters, represented by the matrix $\mathbf{W} \in \mathbb{R}^{s \times C}$, which define a convolution operation across the $m$ nodes of $\mathcal{G}$ (where $m = |\mathcal{V}|$). This convolution is formulated as $(\mathcal{G} \star \mathcal{F})_\mathcal{V} = g\big(\mathbf{A} \mathbf{U}^\top \mathbf{W}\big)$, where $\mathbf{U} \in \mathbb{R}^{s \times m}$ is the graph signal matrix, and $g(\cdot)$ is a pointwise nonlinear activation function. In this operation, the input signal $\mathbf{U}$ undergoes a projection via the adjacency matrix $\mathbf{A}$, effectively aggregating signals from the neighborhood of each node $v$. The elements of $\mathbf{A}$ can be either pre-specified or learned. Consequently, $(\mathcal{G} \star \mathcal{F})_\mathcal{V}$ can be interpreted as a two-layer convolutional block: the first layer aggregates signals from the neighborhood $\mathcal{N}(\mathcal{V})$ of each node $v$ through multiplication of $\mathbf{U}$ by $\mathbf{A}$, while the second layer performs the convolution by applying the $C$ filters in $\mathbf{W}$ to the resulting aggregated signals.
\subsection{Proposed stable bidirectional GCNs}
We formally represent a given GCN as a multi-layered neural network $f$ parameterized by a set of weights $\theta = \{\mathbf{W}_1, \dots, \mathbf{W}_L\}$, where $L$ denotes the network depth. The weight matrix for the $\ell$-th layer is given by $\mathbf{W}_\ell \in \mathbb{R}^{d_{\ell-1} \times d_\ell}$, with $d_\ell$ being the dimensionality of the $\ell$-th layer's output. The output of the $\ell$-th layer, denoted by $\mathbf{\Phi}^\ell$, is defined as $\mathbf{\Phi}^{\ell} = g_\ell(\mathbf{W}_\ell^\top \mathbf{\Phi}^{\ell-1})$ for $\ell \in \{2, \dots, L\}$, where $g_\ell$ is a nonlinear activation function applied element-wise. For notational simplicity, we omit the bias term in the definition of $\mathbf{\Phi}^{\ell}$. \\
\indent In this section, our focus is on designing invertible and stable bidirectional networks. The invertibility (bijection) of a function $f: \mathbb{R}^p \to \mathbb{R}^q$ establishes a {\it one-to-one} correspondence between $\mathbb{R}^p$ and $\mathbb{R}^q$ (necessitating $p = q$)\footnote{Given that the output dimensionality of $f$ depends on the number of classes, a straightforward technique involves augmenting the output space with auxiliary dimensions to match any desired dimensionality, a strategy applicable to other layers as well.}. This property ensures that (i) distinct network inputs, $\mathbf{\Phi}^1_1$ and $\mathbf{\Phi}^2_1$, are mapped to distinct outputs $\mathbf{\Phi}_L^1$, $\mathbf{\Phi}_L^2$, and (ii) for every output $\mathbf{\Phi}_L$, there exists at least one input $\mathbf{\Phi}_1$ such that $f(\mathbf{\Phi}_1) = \mathbf{\Phi}_L$, effectively rendering the trained GCNs bidirectional. Stability extends invertibility by ensuring that the inverse mapping $f^{-1}$, when evaluated on a target latent distribution (e.g., Gaussian), does not diverge significantly from the ambient (input) distribution, and vice versa.
\begin{definition}[Stability]\it 
A bidirectional network $f : \mathbb{R}^p \rightarrow \mathbb{R}^q$ is termed bi-Lipschitzian (or KM-Lipschitzian) if $f$ is $K$-Lipschitz continuous and its inverse $f^{-1}$ is $M$-Lipschitz continuous. The $KM$-Lipschitz constant of such a network is defined as the product $K \times M$.
\end{definition}
Generally, achieving small values for both the Lipschitz constant $K$ and its inverse's Lipschitz constant $M$ for an arbitrary nonlinear function is a non-trivial task \cite{heinonen2005lectures}, consequently making the $KM$ constant small is equally challenging. However, our subsequent bidirectional network design, under specific conditions, enables the $KM$ constant to be small, ideally approaching 1, as demonstrated by our proposition below.
\begin{proposition}\label{prop2}\it 
Given that: (i) the pointwise activation functions $\{g_\ell(\cdot)\}_{\ell=2}^L$ are bijective in $\mathbb{R}^p$; (ii) their derivatives satisfy $l \leq |g'_\ell(\cdot)| \leq u$; and (iii) the condition numbers of the weight matrices in $\theta$ are bounded by $\kappa$, then the bidirectional network $f$ is $KM$-Lipschitz continuous with the $KM$-Lipschitz constant 
\begin{equation}\label{bound0}KM = \left(\kappa \frac{u}{l}\right)^{L-1}.\end{equation}
\end{proposition}
Again, due to space limit, details of the proof are omitted. More importantly, following the aforementioned proposition, when $f$ is invertible in $\mathbb{R}^p$, its inverse $f^{-1}(\mathbf{\Phi}^{L})=\mathbf{\Phi}^{1}$ can be derived, where $\mathbf{\Phi}^{\ell-1}=(\mathbf{W}_\ell^\top)^{-1} g_\ell^{-1}(\mathbf{\Phi}^{\ell})$. The condition number of a matrix $\mathbf{W}_\ell$, defined as $\|\mathbf{W}_\ell\|_2 \|\mathbf{W}_\ell^{-1}\|_2$, quantifies the sensitivity of $\mathbf{W}_\ell$ to small perturbations in $\mathbf{\Phi}^{\ell-1}$ and $\mathbf{\Phi}^\ell$. A small condition number indicates a well-conditioned matrix. When $\kappa$, $l$, and $u$ are close to 1, then $KM \approx 1$, implying that the bidirectional network $f$ is approximately 1-Lipschitz continuous. Consequently, small updates to exemplars in the latent space (via the fixed-point iteration in Eq.~\ref{eq2}) will result in correspondingly small updates to these exemplars in the ambient space when applying $f^{-1}$. This eventually leads to a stable exemplar design in the ambient space, where the exemplars adhere to the actual distribution of the data manifold. \\ 

\noindent As the Lipschitz constant of $f$ is given by $\prod_{\ell} \|\mathbf{W}_\ell \|_2 |g'_\ell|$, and for $f^{-1}$ it is $\prod_{\ell} \|(\mathbf{W}_\ell^\top)^{-1}\|_2 |(g^{-1}_\ell)'|$, the sufficient conditions ensuring that the bidirectional network is $KM$-Lipschitz continuous with a small $KM$ constant are again: (1) small condition numbers $\{\|\mathbf{W}_\ell \|_2 \|\mathbf{W}_\ell^{-1} \|_2\}_\ell$, and (2) $l, u \approx 1$ (with $l < u$ to guarantee the nonlinearity of $f$). By design, conditions (1) and (2) can be satisfied by choosing the slope of the activation functions to be close to one (in practice, $u=0.99$ and $l=0.95$, corresponding respectively to the positive and negative slopes of the leaky-ReLU\footnote{This setting ensures a small ratio between $u$ and $l$, contributing to a small $KM$ constant $(\kappa u/l)^{L-1}$, also dependent on the condition number $\kappa$ (refer again to Proposition~\ref{prop2}).}), and by constraining all weight matrices to have a low condition number. This is achieved by adding a regularization term to the cross-entropy (CE) loss when training GCNs, as:
\begin{equation}\label{eqqqqq2}
\min_{\{\mathbf{W}_\ell\}_\ell}{\textrm{CE}}(f;\{\mathbf{W}_\ell\}_\ell) + \lambda \sum_{\ell} \|\mathbf{W}_\ell \|_2 \|\mathbf{W}_\ell^{-1} \|_2.
\end{equation}
While this formulation is theoretically sound and specifically tailored to our objective (i.e., learning stable bidirectional networks), optimizing the condition number poses a significant challenge due to its non-convexity and non-smoothness, making traditional gradient-based optimization difficult. Furthermore, the condition number's dependence on eigenvalues, as nonlinear measures of matrices, renders gradient estimation unstable and optimization challenging, especially for large-scale matrices. Moreover, balancing cross-entropy and condition number minimization further complicates the problem (see later performance results in Tables~\ref{abla}-\ref{ablb}). Consequently, we consider a surrogate term that {\it formally} achieves optima with unitary condition numbers—analogous to the regularizer in Eq.~\ref{eqqqqq2}—while making optimization more tractable in practice, thereby exhibiting better performance (as demonstrated later in the experiments). Hence, instead of directly minimizing the condition number in the loss, we constrain the matrices in $\theta$ to be {\it orthonormal}, which also guarantees their invertibility. With this modification, the global loss function for training GCNs becomes:
\begin{equation}\label{eqqqqq} 
\min_{\{\mathbf{W}_\ell\}_\ell}{\textrm{CE}}(f;\{\mathbf{W}_\ell\}_\ell) + \lambda \sum_{\ell} \|\mathbf{W}_\ell^\top \mathbf{W}_\ell-\mathbf{I}\|_F,
\end{equation}
where $\mathbf{I}$ denotes identity matrix, $\|\cdot\|_F$ is the Frobenius norm, and $\lambda > 0$ (with $\lambda = \frac{1}{p}$ in practice\footnote{Note that in data-scarce regimes, the cross-entropy term involves few labeled samples, thus setting $\lambda$ to small values is sufficient to ensure minimization of both terms.}). In particular, when $\mathbf{W}_\ell^\top \mathbf{W}_\ell - \mathbf{I} = 0$, then $\mathbf{W}_\ell^{-1} = \mathbf{W}_\ell^\top$ and $\|\mathbf{W}_\ell \|_2 = \|\mathbf{W}_\ell^{-1} \|_2 = 1$, resulting in a tighter $KM$-Lipschitz constant in Eq.~\ref{bound0}. With this updated loss, the learned GCNs are guaranteed to be invertible and stable, while also exhibiting strong discriminative capabilities, as shown later in experiments.
\subsection{Weight reparametrization}
To further enhance the stability of the learned network $f$, we introduce a {\it weight reparametrization} (WR) defined as $\{\mathbf{W}_\ell = \hat{\mathbf{W}}_\ell + \delta \mathbf{I}\}_\ell$, where $\delta \geq 0$ and $\mathbf{I}$ is the identity matrix. This transformation ensures that the eigenvalues of $\mathbf{W}_\ell$, given by $\{\lambda_i + \delta\}_i$ (where $\{\lambda_i\}_i$ are the eigenvalues of $\hat{\mathbf{W}}_\ell$), are bounded below by $\delta$. Consequently, the condition number of $\mathbf{W}_\ell$ is further reduced to $\frac{\max_i |\lambda_i + \delta|}{\min_i |\lambda_i + \delta|}$. A lower condition number implies that small perturbations in latent space exemplars (via Eq.~\ref{eq2}) will result in correspondingly small perturbations in the ambient (input) space upon applying $f^{-1}$, and conversely, small changes in ambient space data will yield stable responses from $f$. While this WR guarantees a minimum eigenvalue of $\delta$, achieving an optimal condition number (close to unity) without excessively increasing $\delta$ and compromising the network's expressiveness remains challenging with this reparametrization alone. Therefore, explicit regularization of the cross-entropy loss, as shown in Eqs.~\ref{eqqqqq2} and \ref{eqqqqq}, is also crucial to avoid the need for overestimated $\delta$ values (see Tables~\ref{abla}-\ref{ablb}). Notably, with this WR, the gradient of the loss in Eqs.~\ref{eqqqqq2}-\ref{eqqqqq} with respect to $\hat{\mathbf{W}}$, denoted as $\nabla_{\hat{\mathbf{W}}} \mathcal{L}$, remains identical to $\nabla_{\mathbf{W}} \mathcal{L}$ since $\nabla_{\hat{\mathbf{W}}} \mathcal{L} = \nabla_{\mathbf{W}} \mathcal{L} \cdot \frac{\partial \mathbf{W}}{\partial \hat{\mathbf{W}}}$ (by the chain rule), and $\frac{\partial \mathbf{W}}{\partial \hat{\mathbf{W}}}$ is simply the identity matrix (as $\mathbf{W} = \hat{\mathbf{W}} + \delta \mathbf{I}$). Thus, this WR directly shifts the eigenvalues, further improving stability without altering the loss gradient.
\section{Experiments}\label{sec:experiments}
This section evaluates the performance of baseline GCNs and our proposed label-frugal GCNs on skeleton-based action recognition using the SBU Interaction \cite{Yun2012} and First Person Hand Action (FPHA) \cite{refref11} datasets. The SBU Interaction dataset, captured with a Microsoft Kinect, contains 282 skeleton sequences of two interacting individuals performing one of eight predefined dyadic actions. Each sequence comprises the 3D spatial coordinates of 15 joints for each person over time. Evaluation follows the standard train-test split defined in \cite{Yun2012}. The FPHA dataset consists of 1175 skeleton sequences spanning 45 diverse hand action categories performed by six subjects in three scenarios, exhibiting substantial intra-class variations in style, speed, scale, and viewpoint. Each sequence represents the temporal evolution of the 3D coordinates of 21 hand joints. Adhering to the evaluation protocol in \cite{refref11}, we employ a 1:1 train-test split, with 600 sequences for training and 575 for testing. For both datasets, we report the average classification accuracy across all action classes.\\
\noindent {\bf Input graphs.} Each skeleton sequence $\{S_t\}_{t=1}^{T}$, consisting of 3D joint coordinates $S_t = \{\hat{p}_{tj}\}_{j=1}^{J}$ over $T$ frames and $J$ joints, is transformed into a graph $\mathcal{G} = (\mathcal{V}, \mathcal{E})$. The nodes $\mathcal{V}$ correspond to joint trajectories $v_j$, where each trajectory $\{\hat{p}_{tj}\}_{t=1}^{T}$ represents the temporal evolution of a joint's 3D position. The edges $\mathcal{E}$ connect trajectories of spatially adjacent joints $(v_j, v_i) \in \mathcal{E}$. To incorporate temporal information, each joint trajectory is divided into $M_c = 4$ equal temporal segments (chunks). Joint coordinates $\{\hat{p}_{tj}\}_{t=1}^{T}$ are assigned to these chunks based on their frame indices. Within each chunk, the mean 3D joint coordinates are computed and concatenated to form a trajectory descriptor $\psi(v_j) \in \mathbb{R}^s$ with dimensionality $s = 3M_c$. This temporal chunking approach encodes temporal dynamics while providing robustness to variations in frame rate and sequence length.

\begin{table}  
\centering
  \begin{minipage}[c]{1\columnwidth}
\centering
    \resizebox{0.56\columnwidth}{!}
{
\begin{tabular}{cc|c}
{\bf Method}      &   & {\bf Accuracy (\%)}\\
\hline 
  Raw Position \cite{Yun2012} & $ \ $   & 49.7   \\ 
  Joint feature \cite{Ji2014}  & $ \ $   & 86.9   \\
  CHARM \cite{Li2015a}       & $ \ $    & 86.9   \\
 \hline  
H-RNN \cite{Du2015}         & $ \ $    & 80.4   \\ 
ST-LSTM \cite{Liu2016}      & $ \ $    & 88.6    \\ 
Co-occurrence-LSTM \cite{Zhua2016} & $ \ $  & 90.4  \\ 
STA-LSTM  \cite{Song2017}     & $ \ $   & 91.5  \\ 
ST-LSTM + Trust Gate \cite{Liu2016} & $ \ $  & 93.3 \\
VA-LSTM \cite{Zhang2017}      & $ \ $  & 97.6  \\
 GCA-LSTM \cite{GCALSTM}                    &   $ \ $      &  94.9     \\ 
  \hline
Riemannian manifold. traj~\cite{RiemannianManifoldTraject} &  $ \ $  & 93.7 \\
DeepGRU  \cite{DeepGRU}        &    $ \ $   &    95.7    \\
RHCN + ACSC + STUFE \cite{Jiang2020} & $ \ $   & 98.7 \\ 
  \hline
\hline 
  \textcolor{black}{Our baseline GCN} &              &        98.4      
\end{tabular}}
\caption{Comparison of our baseline GCN (not label-efficient) against related work on the SBU database.}\label{tab222}
\end{minipage}
\begin{minipage}[c]{0.69\columnwidth}
\resizebox{1\columnwidth}{!}{
\begin{tabular}{ccccc}
{\bf Method} & {\bf Color} & {\bf Depth} & {\bf Pose} & { \bf Accuracy (\%)}\\
\hline
  2-stream-color \cite{refref10}   & \cmark  &  \xmark  & \xmark  &  61.56 \\
 2-stream-flow \cite{refref10}     & \cmark  &  \xmark  & \xmark  &  69.91 \\  
 2-stream-all \cite{refref10}      & \cmark  & \xmark   & \xmark  &  75.30 \\
\hline 
HOG2-dep \cite{refref39}        & \xmark  & \cmark   & \xmark  &  59.83 \\    
HOG2-dep+pose \cite{refref39}   & \xmark  & \cmark   & \cmark  &  66.78 \\ 
HON4D \cite{refref40}               & \xmark  & \cmark   & \xmark  &  70.61 \\ 
Novel View \cite{refref41}          & \xmark  & \cmark   & \xmark  &  69.21  \\ 
\hline
1-layer LSTM \cite{Zhua2016}        & \xmark  & \xmark   & \cmark  &  78.73 \\
2-layer LSTM \cite{Zhua2016}        & \xmark  & \xmark   & \cmark  &  80.14 \\ 
\hline 
Moving Pose \cite{refref59}         & \xmark  & \xmark   & \cmark  &  56.34 \\ 
Lie Group \cite{Vemulapalli2014}    & \xmark  & \xmark   & \cmark  &  82.69 \\ 
HBRNN \cite{Du2015}                & \xmark  & \xmark   & \cmark  &  77.40 \\ 
Gram Matrix \cite{refref61}         & \xmark  & \xmark   & \cmark  &  85.39 \\ 
TF    \cite{refref11}               & \xmark  & \xmark   & \cmark  &  80.69 \\  
\hline 
JOULE-color \cite{refref18}         & \cmark  & \xmark   & \xmark  &  66.78 \\ 
JOULE-depth \cite{refref18}         & \xmark  & \cmark   & \xmark  &  60.17 \\ 
JOULE-pose \cite{refref18}         & \xmark  & \xmark   & \cmark  &  74.60 \\ 
JOULE-all \cite{refref18}           & \cmark  & \cmark   & \cmark  &  78.78 \\ 
\hline 
Huang et al. \cite{Huangcc2017}     & \xmark  & \xmark   & \cmark  &  84.35 \\ 
Huang et al. \cite{ref23}           & \xmark  & \xmark   & \cmark  &  77.57 \\  
\hline 
HAN  \cite{Liu2021}   & \xmark  & \xmark   & \cmark & 85.74 \\
  \hline
  \hline
Our baseline GCN                   & \xmark  & \xmark   & \cmark  & 88.17                                                  
\end{tabular}}
\vspace{0.25cm}
\caption{Same caption as table~\ref{tab222} but for FPHA.}\label{compare2}
\end{minipage}
\end{table}
\noindent {\bf Implementation details \& baseline GCNs.} All GCN models were trained for 2700 epochs using the Adam optimizer with a momentum of 0.9. The batch size was set to 200 for SBU and 600 for FPHA. We employed an adaptive learning rate $\nu$, dynamically adjusted based on the temporal derivative of the loss (Eqs.~\ref{eqqqqq2}-\ref{eqqqqq}): $\nu$ was multiplied by 0.99 upon an increase in the temporal derivative and by 1/0.99 otherwise. Training was performed on a GeForce GTX 1070 GPU with 8 GB memory, without dropout or data augmentation. For SBU, the GCN comprised three sequential blocks, each containing a single-head attention mechanism followed by a convolutional layer with 8 filters, succeeded by a fully connected layer and a classification layer. For the more challenging FPHA dataset, we used a larger GCN, differing primarily in the convolutional layers which employed 16 filters. As detailed in Tables~\ref{tab222} and \ref{compare2}, these baseline GCNs achieved high classification accuracy on both SBU and FPHA. Our subsequent goal is to achieve as close as possible (comparable) performance with significantly fewer labeled samples through label-efficient learning.

\begin{table}[h]
 \begin{center}
\resizebox{0.69\columnwidth}{!}{
  \begin{tabular}{cll}    
   \rotatebox{0}{Labeling rates}  &     \rotatebox{0}{Accuracy}  & \rotatebox{0}{Observation}  \\
 \hline
  \hline
    100\%    &    98.40     & Baseline GCN (not label-efficient)\\
     \hline
    \multirow{5}{*}{\rotatebox{0}{45\%}}     & \underline{89.23}   & wo display model (random display)  \\
                                  &  \underline{89.23}   & + display model + ambient  (our) \\
                                  &   \bf93.84   & + display model + latent  (our) \\
                                  &67.69   & uncertainty (margin-based) \\                                                                                                                                          & 83.07  & diversity (coreset-based)\\                                        
    \hline
      \multirow{5}{*}{\rotatebox{0}{30\%}}   & 80.00  & wo display model (random display)  \\
                                             & \underline{86.15}  & + display model + ambient  (our) \\
                                  & \bf87.69   & + display model + latent  (our) \\                 
                                             & 61.53  & uncertainty (margin-based) \\                                                                                                                                                                              & 83.07   & diversity (coreset-based)\\                                        
    \hline 
        \multirow{5}{*}{\rotatebox{0}{15\%}} & \underline{69.23}  & wo display model (random display) \\
                                             & \bf75.38  & + display model + ambient   (our) \\
                                  & \bf75.38  & + display model + latent  (our) \\
                                             & 56.92  & uncertainty (margin-based) \\                                                                                                                                                                              & 66.15  & diversity (coreset-based) \\                                        
      \hline 
  \end{tabular}}
\end{center}
\caption{This table shows detailed performances and ablation study on SBU for different  labeling rates. Here ``wo'' stands for ``without''. Best results are shown in bold and second best results underlined.}\label{table21}
\end{table}

\subsection{Display model: comparison \& ablation}
Tables~\ref{table21} and \ref{table22} present a comparative analysis and ablation study of our proposed method on SBU and FPHA, respectively. The results demonstrate that applying our display model directly in the ambient space achieves high classification accuracy, often significantly outperforming comparative display selection strategies. Furthermore, leveraging the latent space yields a noticeable additional performance improvement, underscoring the effectiveness of our model and its synergy with latent representations. When compared against alternative display selection strategies integrated with our GCN learning framework—including random sampling, diversity-based selection \cite{zhang2022multi}, and uncertainty-based selection \cite{zhao2023uncertainty}—our method consistently exhibits substantial performance gains across various equivalent labeling rates. As shown in Tables~\ref{table21} and \ref{table22}, our approach offers significant advantages, particularly in data-scarce scenarios. While random sampling shows competitive performance at higher labeling rates (e.g., 45\%), consistent with previous findings (e.g., \cite{Burr2009}), its effectiveness diminishes considerably at lower rates (e.g., 15\%), necessitating more sophisticated selection techniques. Uncertainty-based selection, while improving classification confidence, lacks sufficient diversity in the selected samples. Conversely, random and diversity-based methods do not adequately refine the classification process. Moreover, all comparative methods are limited by their reliance on selecting displays from a static pool. In contrast, our display model learns adaptable exemplars within the latent space of our stable and invertible bidirectional GCNs, leading to enhanced performance, especially under frugal labeling. This adaptability allows for a more effective data representation, resulting in improved classification accuracy.
\begin{table}[h]
 \begin{center}
\resizebox{0.69\columnwidth}{!}{
  \begin{tabular}{cll}    
   \rotatebox{0}{Labeling rates}  &     \rotatebox{0}{Accuracy}  & \rotatebox{0}{Observation}  \\
 \hline
  \hline
    100\%    &    88.17     & Baseline GCN (not label-efficient)\\
     \hline
    \multirow{7}{*}{\rotatebox{0}{45\%}}     & \underline{75.47} & wo display model (random display)  \\
                                             & 72.52  & + display model + ambient  (our)  \\
                                  & \bf75.65 & + display model + latent   (our) \\
                                                                                    & 63.30   & uncertainty (margin-based) \\                                                                                                                                                                            & 70.26   & diversity (coreset-based)\\         
                                  
    \hline
      \multirow{7}{*}{\rotatebox{0}{30\%}}   & \bf67.47   & wo display model (random display)  \\
                                             & 61.21   & + display model + ambient  (our) \\
                                  & \underline{63.65}   & + display model  + latent  (our) \\

                                             & 56.17   & uncertainty (margin-based) \\                                                                                                                                                                             & 62.08    & diversity (coreset-based)\\                                        
    \hline 
        \multirow{7}{*}{\rotatebox{0}{15\%}} &  40.52   & wo display model (random display)  \\
                                             &  45.21        & + display model + ambient  (our)  \\
                                  & \bf49.21    & + display model + latent  (our) \\
                                
                                             & 41.73    & uncertainty (margin-based) \\                                                                                                                                                                             & \underline{46.26}   & diversity (coreset-based) \\                                       
      \hline 
  \end{tabular}}
\end{center}
\caption{Same caption as table~\ref{table21} but for FPHA.}\label{table22}
\end{table}
\subsection{Regularization and weight reparametrization} 
Tables~\ref{abla} and \ref{ablb} detail the individual and combined effects of our regularization strategies — Condition Number (CN) regularization and Orthogonality Regularization (OR) — alongside  Weight Reparametrization (WR). The results consistently highlight the beneficial impact of WR, both independently and when integrated with regularization. Notably, except for OR regularization alone (config. \#7, \#8), WR significantly reduces both the observed Condition Number (CN) and Fr\'echet Inception Distance (FID), particularly with sufficiently large $\delta$, while simultaneously improving classification accuracy compared to the non-reparametrized baseline (config. \#2, \#3, \#4 vs \#1, and \#6 vs \#5) across various $\delta$ values. An excessively large $\delta$ (config. \#2) introduces excessive rigidity, leading to minimal FID and CN but hindering cross-entropy minimization and thus reducing classification accuracy. Conversely, an underestimated $\delta$ (config. \#4) provides greater model flexibility, facilitating cross-entropy minimization but resulting in poorer generalization, indicated by higher FID and CN scores suggesting out-of-distribution exemplars. An intermediate $\delta$ (config. \#3) strikes a better balance, optimizing the reparametrization's effectiveness. When combined with CN regularization (config. \#6), WR exhibits less dependence on large $\delta$ values and effectively mitigates FID and CN, reducing the sensitivity to precise $\delta$ tuning at higher values, thereby simplifying its selection. Across all experiments, OR (config. \#7, \#8) consistently yields notable improvements in accuracy, FID, and observed CN, both with and without reparametrization, confirming its efficacy as a stronger regularizer compared to CN regularization.
\begin{table}
\centering
  \resizebox{0.85\columnwidth}{!}{
 \begin{tabular}{cc||cccc}
Regularizer              & WR (${\bf W}+\delta I$)  & Acc $\uparrow$ & Observed CN  $\downarrow$& FID Score $\downarrow$ & config\\
   \hline
No                     & No   &  9.23   & $1.85 \times 10^{29}$  &  $6.44 \times 10^{15}$   & \#1\\
No                    & Yes,  $\delta=10^6$ &  58.46   &  2.022 &  \bf7.16   & \#2 \\
No                     &  Yes,  $\delta=10^5$ &  \underline{83.07}   &  154.52  &   8.88   & \#3 \\
   No                     &  Yes,  $\delta=10^1$ &  \underline{83.07}   &  $5.01 \times 10^{11}$  & 92.04   & \#4 \\
\hline 
CN     & No   & 9.23    &  $3 \times 10^9$  &  3973.2   & \#5 \\
CN     & Yes,  $\delta=10^1$  &  44.23  &  \underline{1.015}  & 15.85    & \#6 \\
\hline    
OR           & No   &  \bf93.84  &  5.410  &  10.18    & \#7  \\
  OR           &  Yes,  $\delta=10^1$  &  81.53  &  \bf1.010 &  \underline{8.70}  & \#8    \\   
 \hline
\end{tabular}}
\vspace{0.02cm}
\caption{\textcolor{black}{This table shows the impact of different regularizers (OR and CN) and WR (for different setting of $\delta$) when taken individually and combined. Here Acc (accuracy), observed CN and FID scores are shown on the SBU dataset.  Best results are shown in bold and second best results underlined.}}\label{abla} 
 \end{table} 

 \begin{table}
\centering
   \resizebox{0.85\columnwidth}{!}{
 \begin{tabular}{cc||cccc}
Regularizer              & WR (${\bf W}+\delta I$)   & Acc $\uparrow$ & Observed CN  $\downarrow$& FID Score $\downarrow$ & config\\ 
\hline
No                     & No   & 54.78    &  $2.91 \times 10^{22}$  &  $5.30 \times 10^{9}$   & \#1 \\
No                     &  Yes,  $\delta=10^{6}$ &  2.26  &  4.666  &  6.32   & \#2 \\
No                     &  Yes,  $\delta=10^{5}$ &   54.78   &  32.362  &  5.87    & \#3 \\
No                     &  Yes,  $\delta=10^1$ &  57.04   & $1.19 \times 10^{11}$ &  13.33   & \#4 \\
\hline 
CN      & No   & 2.08   &   $2.89 \times 10^{30}$  & $1.86 \times 10^{12}$   & \#5 \\
CN   &  Yes,  $\delta=10^1$  & 64.17  & \bf1.000  &  7.05   & \#6 \\
\hline   
OR            & No   &  \bf75.65  &  \underline{1.052}  & \bf2.37    & \#7 \\
OR            &  Yes,  $\delta=10^1$  &  \underline{68.34}  &  1.055  & \underline{5.54}    & \#8    \\
 \hline
\end{tabular}}
\vspace{0.02cm}
\caption{\textcolor{black}{Same caption as table~\ref{abla} but for FPHA.}}\label{ablb} 
 \end{table} 
\section{Conclusion} 
This paper presents a label-efficient approach for skeleton-based action recognition leveraging graph convolutional networks (GCNs), significantly reducing the reliance on extensive labeled data and thereby enhancing the applicability of GCNs in annotation-scarce settings. The core contribution of this work is the formulation of a novel acquisition function, derived as the solution to a carefully constructed objective function. This function balances representativeness, diversity, and uncertainty to yield a selection of unlabeled data that optimally capture the underlying distribution. Moreover, we further refine our framework by developing bidirectional and stable GCNs, resulting in learned latent spaces with improved representational fidelity and discriminative capacity. Comprehensive experiments conducted on two challenging skeleton-based action recognition datasets  validate the effectiveness and superior performance of our proposed method.


\begin{thebibliography}{1}
\bibitem{aaaaa9}  H. Sahbi.  A particular Gaussian mixture model for clustering and its application to image retrieval.   Soft Computing 12 (7), 667-676
\end{thebibliography}
\end{document}